\documentclass[10pt,twocolumn,letterpaper]{article}

\usepackage{cvpr}              

\usepackage[dvipsnames]{xcolor}

\definecolor{cvprblue}{rgb}{0.21,0.49,0.74}
\usepackage[pagebackref,breaklinks,colorlinks,citecolor=cvprblue]{hyperref}
\usepackage{multirow}
\usepackage{booktabs}
\usepackage{bm}
\usepackage{float}

\usepackage{siunitx}
\usepackage{array}

\def\paperID{6} 
\def\confName{CVPR}
\def\confYear{2024}

\title{Vision-language models for decoding provider \\ attention during neonatal resuscitation}

\author{
Felipe Parodi\textsuperscript{1}, Jordan K. Matelsky\textsuperscript{2,8}, Alejandra Regla-Vargas\textsuperscript{3},\\
Elizabeth E. Foglia\textsuperscript{6,7}, Charis Lim\textsuperscript{6,7}, Danielle Weinberg\textsuperscript{6,7},\\
Konrad P. Kording\textsuperscript{1,2}, Heidi M. Herrick\textsuperscript{6,7,$\dagger$}, Michael L. Platt\textsuperscript{1,4,5,$\dagger$}\\
\\
\small \textsuperscript{1}Department of Neuroscience, \textsuperscript{2}Department of Bioengineering, \textsuperscript{3}Department of Sociology,\\
\small \textsuperscript{4}Department of Marketing, \textsuperscript{5}Department of Psychology, University of Pennsylvania,\\
\small \textsuperscript{6}Division of Neonatology, Department of Pediatrics, University of Pennsylvania Perelman School of Medicine,\\
\small \textsuperscript{7}Division of Neonatology, Department of Pediatrics, Children's Hospital of Philadelphia,\\
\small \textsuperscript{8}Johns Hopkins University Applied Physics Laboratory\\
\small Correspondence: \texttt{herrickh@chop.edu}; \texttt{fparodi@pennmedicine.upenn.edu}
}

\begin{document}
\maketitle

\def\thefootnote{\fnsymbol{footnote}}

\footnotetext[2]{Heidi Herrick and Michael Platt contributed equally to this work.}

\setlength{\abovecaptionskip}{5pt}
\setlength{\belowcaptionskip}{-5pt}
\begin{figure*}
    \centering
\includegraphics[width=0.8\textwidth]{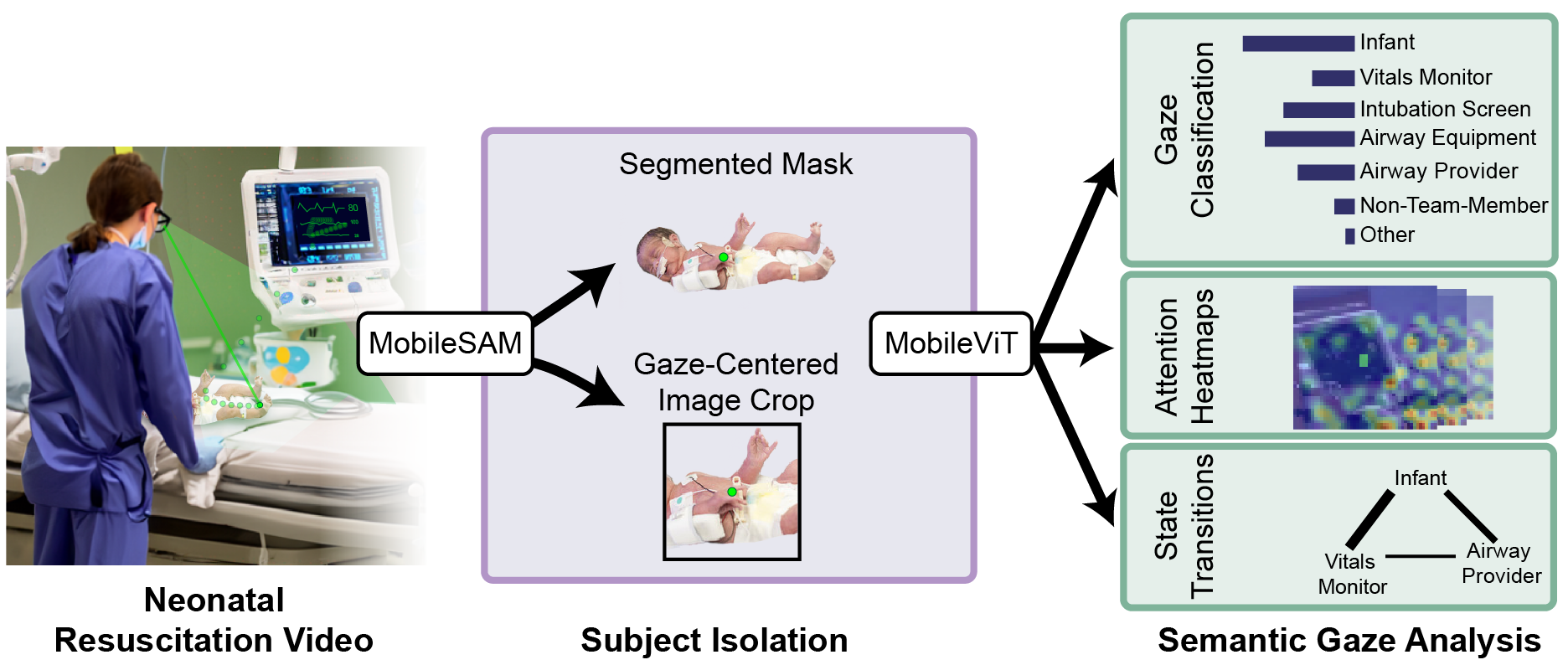}
    \captionsetup{skip=5pt}
    \caption{\textbf{Approach.} (Left) During resuscitation, physicians must attend to multiple stimuli at once. (Middle) The output of Tobii eye-tracking glasses can be used to isolate the subject with segmentation (top) and cropping (bottom). (Right) Cropped images and object masks are then fed to the model for semantic gaze classification, and prediction scores are aggregated for each target for attention analysis. Note: depicted infant is synthetic.}
    \label{fig:fig1}
\end{figure*}

\begin{abstract}
Neonatal resuscitations demand an exceptional level of attentiveness from providers, who must process multiple streams of information simultaneously. Gaze strongly influences decision making; thus, understanding where a provider is looking during neonatal resuscitations could inform provider training, enhance real-time decision support, and improve the design of delivery rooms and neonatal intensive care units (NICUs). Current approaches to quantifying neonatal providers' gaze rely on manual coding or simulations, which limit scalability and utility. Here, we introduce an automated, real-time, deep learning approach capable of decoding provider gaze into semantic classes directly from first-person point-of-view videos recorded during live resuscitations. Combining state-of-the-art, real-time segmentation with vision-language models (CLIP), our low-shot pipeline attains 91\% classification accuracy in identifying gaze targets without training. Upon fine-tuning, the performance of our gaze-guided vision transformer exceeds 98\% accuracy in gaze classification, approaching human-level precision. This system, capable of real-time inference, enables objective quantification of provider attention dynamics during live neonatal resuscitation. Our approach offers a scalable solution that seamlessly integrates with existing infrastructure for data-scarce gaze analysis, thereby offering new opportunities for understanding and refining clinical decision making.
\end{abstract}    
\section{Introduction}
\label{sec:intro}


Neonatal resuscitation is a complex process in which a lead provider is tasked with overseeing the resuscitation progression, monitoring vital signs, and coordinating team response, often within the confines of a bustling delivery room \citep{herrick_impact_2020, Weiner_Zaichkin, weinberg_visual_2020}. Even a momentary attentional lapse can escalate the risk of errors and adverse outcomes \citep{yamada_analysis_2015}, making it imperative to identify sources of inefficiency and care disruptions \citep{herrick_impact_2020}. Quantitative assessment of visual attention not only aids in pinpointing sources of inefficiency, but also advances patient care, improves training protocols for medical practitioners \citep{leone_using_2019, williams_can_2013, wilson_gaze_2011}, and bolsters real-time decision support \citep{szulewski_combining_2014, visweswaran_evaluation_2021}.


Traditionally, monitoring provider visual attention during resuscitation relied heavily on manual annotations of egocentric videos, primarily captured via head-mounted eye-tracking systems. These tools have been deployed in both simulated \citep{damji_analysis_2019, garvey_simulation_2020, mcnaughten_clinicians_2018, szulewski_combining_2014, katz_visual_2019} and real-world settings \citep{law_analysis_2018, zehnder_using_2021, weinberg_visual_2020, herrick_provider_2020}. While valuable insights have been gained through these approaches, including discernible patterns of visual attention on infants, monitors, and team members \citep{wagner_visual_2022, weinberg_visual_2020}, the manual data extraction is time intensive and not scalable. Integration of eye-tracking glasses has partially addressed these gaps, offering a glimpse into physician gaze by providing the location for object fixations and saccades during a given session \citep{law_analysis_2018, wagner_eye-tracking_2020} or for evaluating factors in accessing neonatal equipment \citep{chen_use_2023}. Prior work demonstrates multiple links relating visual search patterns and levels of expertise, including dwell times, eye movements, and search patterns \citep{weinberg_visual_2020, castner_deep_2020, mcnaughten_clinicians_2018, van_der_gijp_how_2017}. Developing a fast, robust, automated system capable of performing semantic gaze analysis is, therefore, a priority \citep{wolf_automating_2018, stubbemann_neural_2021}.

Such a semantic gaze analysis system should decipher the natural language labels associated with a provider's gaze during complex video scenes, even in situations where data availability is restricted due to privacy considerations in medical settings. Automated gaze analysis would enhance medical education and healthcare, fostering optimal attention strategies and improving the efficacy of neonatal resuscitations by providing nuanced feedback on gaze patterns, optimal team configurations, and task allocations.

Here, we introduce a real-time, data-driven pipeline that automates the analysis of provider visual attention patterns during neonatal resuscitations. Our system first isolates objects of interest using real-time instance segmentation from MobileSAM \citep{mobile_sam} and cropping, which are then jointly classified into various semantic labels by a vision transformer, including MobileViT \citep{mehta_mobilevit_2022} and CLIP \citep{radford_learning_2021}, with a top-3 accuracy – the percentage of samples for which the true label is among the top three predicted labels – reaching 98\% (chance = 14\%). Our pipeline, trained on a novel egocentric NICU dataset, integrates outputs from commercial eye trackers and can operate in real time. This approach, validated against human experts, enables an unprecedented level of precision and efficiency in identifying gaze targets among clinically significant regions of interest (ROIs).

\begin{table*}[htp]
\centering
\caption{Breakdown of the Egocentric Dataset of Infant Resuscitation by physician expertise and image distribution by class.}
\label{tab:table1}
\setlength{\tabcolsep}{5.4pt}
\begin{tabular}{p{1.9cm} p{1.4cm} p{1.6cm} p{1.2cm} p{1.2cm} p{1.2cm} p{1.2cm} p{1.3cm} p{1.1cm} p{1.1cm}}
  \toprule 
  \textbf{Physician Recording} & \textbf{Length (min:sec)} & \textbf{Train:Val} & \textbf{Airway Equip.} & \textbf{Airway Prov.} & \textbf{Laryng. Screen} & \textbf{Infant} & \textbf{Vitals Monitor} & \textbf{Non Team} & \textbf{Other} \\
  \midrule 
  Attending\_1 & 02:19 & 2,379: 567 & 440 & 171 & 382 & 592 & 815 & 206 & 340 \\
  Fellow\_8 & 01:02 & 1,088: 298 & 64 & 21 & 0 & 799 & 498 & 0 & 4 \\
  Attending\_26 & 01:28 & 1,175: 274 & 313 & 2 & 356 & 636  & 106 & 0 & 36 \\
  Attending\_29 & 02:11 & 2,424: 603 & 446 & 141 & 539 & 1,474 & 197 & 16 & 214 \\
  Fellow\_30 & 01:09 & 1,293: 326 & 95 & 36 & 1164 & 149  & 153 & 0 & 22 \\
  Attending\_44 & 03:31 & 3,387: 873 & 840 & 347 & 503 & 2,170  & 115 & 39 & 246\\
  Attending\_31 & 04:01 & \multicolumn{8}{l}{Held out for analysis} \\
  Fellow\_56 & 01:00 & \multicolumn{8}{l}{Held out for analysis} \\
  Fellow\_62 & 05:51 & \multicolumn{8}{l}{Held out for analysis} \\
  \bottomrule 
\end{tabular}
\end{table*}

\section{Related Work}

\label{sec:relatedwork}

\textbf{Deep learning applications in healthcare.} Recent years have seen a proliferation of deep learning applications in healthcare. These include deployment of segmentation in radiology to isolate organs of interest from X-ray images \citep{khosravan_collaborative_2019, lee_identification_2022, wang_gazegnn_2023}, use of image classification algorithms to categorize diseases \citep{karargyris_creation_2021}, and implementation of gaze estimation in the operating room to understand surgical decision making \citep{kulkarni_scene-dependent_2023}. Despite these strides, characterizing physician gaze through deep learning remains challenging, constrained by the limited availability of annotated datasets and by the lack of effective, low-shot models. Recent studies have highlighted the utility of zero-shot vision-language models like CLIP in clinical settings \citep{radford_learning_2021, agarwal_evaluating_2021, mishra_improving_2023}, endorsing the potential of heavily pre-trained deep learning models for gaze analysis in data-scarce healthcare settings.

Among emerging tools in deep learning for health, Vision Transformers (ViT) have ushered in a new era of medical analysis \citep{dosovitskiy_image_2021}. Various ViT variants have been developed, focusing on enhancing generalizability, reducing latency, and improving data-cost-effectiveness \citep{dosovitskiy_image_2021, mehta_mobilevit_2022, touvron_going_2021} in data-austere environments, such as in predicting COVID-19 from chest X-ray images \citep{park_vision_2021}. The promise of these models invites further research extending them to real-world settings.

\textbf{Gaze tracking in healthcare.} Gaze tracking technology has permeated the healthcare sector, augmenting medical image interpretation and enhancing the diagnostic, treatment, and monitoring processes by providers \citep{lee_identification_2022, visweswaran_evaluation_2021, wei_machine_2023, zehnder_using_2021}. This technology has proven useful in high-risk domains such as childbirth and neonatal resuscitation, offering critical insights into the interactions between individuals and their surroundings. There has also been a surge in the adoption of semi-automated provider gaze tracking technology, integrating eye-tracking glasses and multi-modal approaches to quantify providers’ attention during medical procedures \citep{stember_integrating_2021,stember_eye_2019}. These techniques, however, have not yet been extended to real-time operation or semantic gaze analysis. Addressing these limitations, recent efforts have sought to incorporate eye-tracking data into deep learning models to provide an interpretable analysis of visual attention patterns \citep{khosravan_collaborative_2019,ma_eye-gaze-guided_2022,stember_eye_2019, wang_gazesam_2023, wang_follow_2022}. Despite these promising trends, the transition from controlled experiments, including simulations, to real-world clinical settings remains a formidable challenge.

\textbf{Semantic gaze analysis.} Semantic gaze analysis decodes the objects of gaze fixations into natural language, offering a deeper understanding of observer intent, situational awareness, and high-level decision-making processes. Despite burgeoning interest in this domain, current studies focus on analyzing gaze patterns in simulated environments \citep{mcnaughten_clinicians_2018, xu_predicting_2014}. A significant gap persists in extending these analyses to real-world clinical settings, especially in high-stakes environments like the delivery room and neonatal intensive care unit.
\section{Approach}
\textbf{Overview.} Here, we introduce a novel framework that integrates in-situ eye-tracking with state-of-the-art neural networks to generate human-interpretable labels for the target of a physician’s gaze during neonatal resuscitation (see Fig.~\ref{fig:fig1} (Left) for a depiction of the eye-tracking gaze estimate). Leveraging the eye-gaze estimate, our pragmatic approach enables real-time gaze characterization in active clinical settings, extending the boundaries of automated and accurate gaze analysis.


\begin{figure*}[htp]
    \centering
    \includegraphics[width=.8\textwidth]{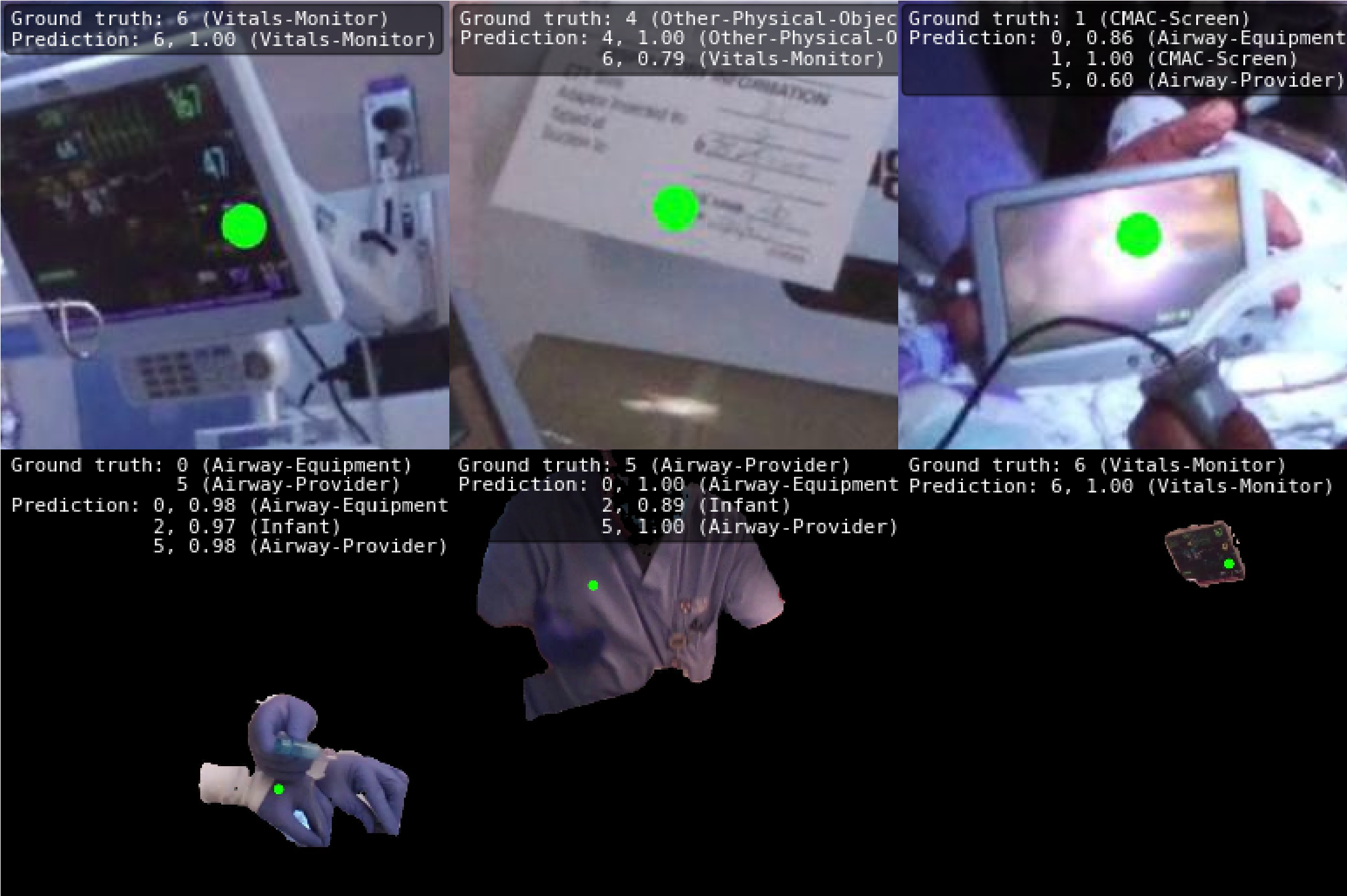}
    \caption{\textbf{Sample gaze classification predictions} on cropped (top) and segmented (bottom) testing images. Note: “CMAC-Screen” refers to "Video Laryngoscope Screen."}
    \label{fig:fig2}
\end{figure*}

\textbf{\textbf{Egocentric Dataset of Infant Resuscitation}.} Nine neonatal resuscitation sessions were recorded using Tobii Pro eye-tracking glasses (Tobii Pro, Stockholm, Sweden), which captured video at a 1080p resolution and 25 frames per second (FPS). These glasses are equipped with gaze-estimation sensors, and the hardware assigns an estimated focus point in each frame. Six videos were used to generate the image dataset and three were held out for analysis. The image dataset consists of expert-annotated segmented and cropped frames, which served as our benchmark in evaluating the efficacy of zero-shot, few-shot, and fine-tuned gaze classification models in learning image features of first-person gaze. Before data collection, eye-tracking calibration was performed for each wearer to ensure accurate gaze estimation. The University Institutional Review Board approved this study, and informed consent was obtained from study participants. The video dataset analyzed in this study cannot be openly released due to privacy considerations. However, the details regarding dataset characteristics (Table \ref{tab:table1}), pre-processing, and model training are provided to maximize the reproducibility of the approach.

\textbf{Annotations.} Before cropping and segmentation, the videos were labeled by annotators using either the Tobii coding software or the DeepEthogram tool \citep{bohnslav_deepethogram_2021}. These annotators consisted of one expert neonatologist with experience in neonatal resuscitation and two graduate students trained to identify regions of interest. Inter-rater reliability was quantified using Cohen’s Kappa, yielding a coefficient of 0.92, indicating substantial agreement among annotators. Annotators labeled the frame into one or more of seven distinct categories: \textit{Infant}, \textit{Vitals Monitor}, \textit{Video Laryngoscope} \textit{Screen}, \textit{Airway Equipment}, \textit{Airway Provider}, \textit{Non-Team Member}, and \textit{Other Physical Objects}. For complex scenes, such as infant intubation, annotators were instructed to consider a two-inch radius in real-world dimensions around the gaze dot. This method relied on the annotators' judgment to infer this radius from the video frame's context to offer a precise portrayal of the physician's gaze trajectory. The annotated image dataset was then split 80:20, resulting in 11,746 training frames (not including their segmented pairs), and 2,941 testing frames (no segmented pairs). This split was frame-centric, ensuring that frames from each video were distributed across both training and testing sets.

\textbf{Identifying optimal input resolution.} We tested several input types to identify the most effective input for zero-shot gaze classification given ground truth annotations: the raw frame (“Frame” in Table \ref{tab:table2}), a 128x128 pixel crop centered around the Tobii gaze estimate (“Crop\textsubscript{128}”), a 256x256 pixel crop around the same estimate (“Crop\textsubscript{256}”), and object pixel masks (“Mask”) for each target class. We used OpenCV to locate the pixel exhibiting the peak green intensity, which then served as the center for the square crop \citep{opencv} and MobileSAM \citep {mobile_sam} to generate the segmented mask given the Tobii gaze estimate. We selected the variants – specifically the “Mask” and “Crop\textsubscript{128}” options – that maximized performance in zero-shot gaze classification to establish the final crop parameters for EDIR. This dataset comprises 14,687 unique frames (crop-segmented pairs), capturing first-person physician perspectives during neonatal resuscitation in the NICU (Fig. \ref{fig:fig1}). The EDIR dataset serves as a critical resource for model deployment in an authentic clinical context.

\textbf{Instance Segmentation with MobileSAM.} To achieve real-time, accurate object segmentation, we utilized MobileSAM, a model recognized for its low latency and robust performance \citep {mobile_sam}. MobileSAM offers the flexibility to use various inputs for mask generation, including bounding boxes, text, or even pixel coordinates. We used an untrained MobileSAM model to ensure that our gaze-classification pipeline was independent of segmentation accuracy. While this decision simplified our pipeline, it did introduce variability in the quality of the pixel masks, potentially affecting the model’s ability to learn specific semantic labels accurately. During segmentation, MobileSAM generated a segmented object mask using the Tobii estimate as input (see Fig. \ref{fig:fig1} (middle) or Fig. \ref{fig:fig2} (bottom) for example masks). These masks delineate the physician's focus for a given frame, isolating the region for subsequent analysis.

\begin{table*}[htbp]
  \centering
  \caption{Semantic gaze prediction under training-free conditions.}
  \label{tab:table2}
  \begin{tabular}{lcccc}
  \toprule
\textbf{Model} & \textbf{Classification} &  \textbf{Input} & \textbf{Top-1 Acc (\%)} & \textbf{Top-3 Acc (\%)}  \\
 \midrule
CLIP-ViT-B-32 & Zero-Shot &  Frame & 8.96 &  38.39 \\
CLIP-ViT-B-32 & Zero-Shot &  Crop$_{128}$ & 36.93 &  62.22 \\
CLIP-ViT-B-32 & Zero-Shot &  Crop$_{256}$ & 37.92 &  49.39 \\
CLIP-ViT-B-32 & Zero-Shot &  Mask & 23.15 &  53.67 \\
\textbf{CLIP-ViT-B-32} & \textbf{Zero-Shot} &  \textbf{Crop$_{\bm{128}}$ + Mask} & \textbf{37.92} &  \textbf{76.10} \\
CLIP-ViT-B-32 & Zero-Shot &  Frame + Crop$_{128}$ + Mask & 37.92 &  56.45 \\
 \midrule
Tip-Adapted-CLIP & \textbf{Few-Shot} &  \textbf{Crop$_{\bm{128}}$} & \textbf{71.17} &  \textbf{91.67} \\
Tip-Adapted-CLIP & Few-Shot &  Crop$_{128}$ + Mask & 54.55 &  84.31 \\
  \bottomrule
\end{tabular}
\end{table*}

\begin{table}[b]
\centering
  \caption{Semantic gaze prediction following supervised training for single- and multi-label classification.}
  \label{tab:table3}
  \setlength{\tabcolsep}{5.5pt}
  \begin{tabular}{lccc}
  \toprule
\textbf{Model} &  \textbf{Input} & \multicolumn{2}{c}{\textbf{Accuracy (\%)}} \\
  \midrule
  & \textit{Single-Label} & \textit{Top-1} & \textit{Top-3} \\
\midrule
ResNet50 &  Crop$_{128}$ + Mask & 81.60 &  96.62 \\
\textbf{MobileViT} &  \textbf{Crop$_{\bm{128}}$ + Mask} & \textbf{93.02} &  \textbf{98.74} \\
CLIP-ViT-B-32 &  Crop$_{128}$ + Mask & 87.54 &  97.19 \\
\midrule
& \textit{Multi-Label} & \textit{mAP} & \textit{F1-Score} \\
\midrule
ResNet50 &  Crop$_{128}$ + Mask & 87.72 &  77.68 \\
\textbf{MobileViT} & \textbf{Crop$_{\bm{128}}$ + Mask} & \textbf{96.71} &  \textbf{91.60} \\
CLIP-ViT-B-32 &  Crop$_{128}$ + Mask & 92.39 &  85.70 \\
  \bottomrule
\end{tabular}
\end{table}

\textbf{Low-Shot Semantic Gaze Classification.} To address the challenge of gaze classification in the data-scarce neonatal intensive care unit, we leveraged the CLIP (Contrastive Language-Image Pre-training) model \citep{radford_learning_2021}, a vision-language model adept at aligning image-text representations, making it effective for zero-shot classification. We employed the ‘base’ vision transformer architecture with 32x32 patches (CLIP-ViT-B-32), which was pre-trained on the LAION-400M image-text dataset \citep{schuhmann_laion-400m_2021}. When performing zero-shot classification, provided class labels are embedded using CLIP's heavily pre-trained text encoder, and the similarity between image and text embeddings is computed, ultimately "predicting" the class exhibiting the highest similarity score. We tested the CLIP-ViT-B-32's zero-shot gaze classification capability on our EDIR at different resolutions: the entire frame, a 128x128 pixel crop centered around the Tobii gaze estimate, a 256x256 pixel crop, and the segmentation mask produced with the Tobii gaze estimate as input to MobileSAM.

Following zero-shot gaze classification, we tested how well CLIP could accurately classify a given cropped or segmented frame under low-shot conditions, in which the model would see only a small set of images from the training set and then perform inference. To do this, we relied on Tip-Adapter \citep{tip_adapter}, which enhances CLIP's few-shot ability by creating a feature adapter from a few-shot (16-image) training set to update CLIP's prior encoded knowledge. Akin to our zero-shot experiments, we tested CLIP's low-shot performance on the 128x128 pixel crop, and on the joint crop-mask pair (Table \ref{tab:table2}).

We evaluated CLIP's performance with Top-1 and Top-3 accuracy on the EDIR testing set (n=2,941). Top-1 accuracy represents the proportion of instances where the true label matches the highest predicted label, whereas Top-3 accuracy accounts for cases where the true label is within the top three predicted labels. This zero- and few-shot paradigm facilitates semantic gaze target prediction without necessitating training on labeled EDIR images, instead harnessing CLIP's generalized knowledge. In subsequent sections, we fine-tune CLIP on EDIR to enhance gaze classification. However, this low-shot evaluation serves as a compelling baseline, showcasing the potential of multimodal representation learning in data-scarce environments.

\textbf{Fine-Tuned Semantic Gaze Classification.}
We next fine-tuned a set of models on our image dataset EDIR. Given that zero-shot gaze classification performance was strongest when combining predictions from both the 128x128 pixel crop and segmentation mask inputs, we opted to use this dual input approach for all subsequent few-shot training experiments given an input image size 224x224 px. This allowed us to leverage the strengths of both localized cropping and precise object masking while maintaining consistency across conditions to enable fair comparison between zero-shot and few-shot settings. For fine-tuned gaze classification, we trained three models for both single-label classification, in which there is only one ground truth label per image, and multi-label classification, in which there may be multiple labels per image.

\textit{Single-Label Gaze Classification.} We trained ResNet50 \citep{he_deep_2016}, MobileViT \citep{mehta_mobilevit_2022}, and CLIP-ViT-B-32 \citep{radford_learning_2021} using the mmpretrain library \citep{mmpretrain_contributors_openmmlabs_2023}. For this task, we incorporated several data augmentations into our training set, including horizontal, vertical, and diagonal flipping. We chose the ResNet-50 as a convolutional baseline and chose the MobileViT model due to its lightweight yet robust performance in diverse computer vision tasks \citep{dosovitskiy_image_2021, touvron_going_2021, mehta_mobilevit_2022}. The ResNet-50 and MobileViT were trained using Stochastic Gradient Descent (SGD), with a learning rate of 0.1, momentum of 0.9, and a weight decay of 0.0001 with a MultiStep Learning Rate Scheduler modulated the learning rate. The CLIP model was trained using a linear and then cosine annealing learning rate. For single-label gaze classification, we minimized cross-entropy loss and evaluated the models using Top-1 and Top-3 accuracy (Table \ref{tab:table4}).

\begin{table*}[htbp]
\caption{Model training and evaluation.}
\label{tab:table4}
\begin{tabular}{ccccccccc}
  \toprule
  Task & Model & Pretraining & Batch Size & Epochs & Loss & Evaluation \\
  \midrule
  Single-label & ResNet-50 & ImageNet-1k \citep{russakovsky_imagenet_2015} & 32 & 100 &  Cross-Entropy & Top-1, Top-3\\
    Single-label & MobileViT (s) & ImageNet-1k  & 128 & 100 &  Cross-Entropy & Top-1, Top-3 \\
      Single-label & CLIP-ViT-B-32 & LAION-400M \citep{schuhmann_laion-400m_2021} & 128 & 300 & Cross-Entropy & Top-1, Top-3 \\
        Multi-label & ResNet-50 & ImageNet-1k & 32 & 100 &  Binary Cross-Entropy & mAP, F1 \\
    Multi-label & MobileViT (s) & ImageNet-1k & 128 & 100 &  Binary Cross-Entropy &  mAP, F1 \\
      Multi-label & CLIP-ViT-B-32 & LAION-400M& 128 & 300 & Binary Cross-Entropy &  mAP, F1  \\
  \bottomrule
\end{tabular}
\end{table*}

\textit{Multi-label Gaze Classification.} We next trained the same set of models – ResNet50, MobileViT, and CLIP-ViT-B-32 – on multi-label gaze classification, in which there can be one or more ground truth labels per image. In complex scenes, such as during infant intubation, the neonatologist must navigate multiple stimuli at once, in which case there may be several areas of interest. For multi-label classification, we minimized the binary cross-entropy loss. We held training conditions constant for ResNet-50 and MobileViT, except for the change of loss function and number of ground truth labels per image. For the multi-label classification task, the traditional vision transformer classification head was replaced with a Multi-Label Linear Classification head with a sigmoid activation function and was trained using the AdamW optimizer. Following training, each model was evaluated using Mean average precision (mAP) and F1-score. Mean average precision (mAP) calculates precision and recall over varying thresholds, balancing the impact of false positives and negatives: a higher mAP indicates better overall performance. The F1-score is the harmonic mean of precision and recall, and offers an insight into the balance achieved between the two, especially vital when dealing with class imbalances, as tends to be the case in data-scarce environments.

\begin{table}[bp]
\caption{Average inference speed (frames per second) for batch sizes 1 and 8 across hardware, rounded to the nearest integer.}
\label{tab:table5}
  \setlength{\tabcolsep}{4.5pt}
\begin{tabular}{llccc}
  \toprule
  & \multicolumn{1}{c}{\textbf{Model}} &  \multicolumn{1}{c}{\textbf{A6000}} & \multicolumn{1}{c}{\textbf{RTX 3080}} & \multicolumn{1}{c}{\textbf{i7-11700K}} \\
  \midrule
  \multirow{3}{*}{\rotatebox{90}{BS 1}} & MobileViT & 138 FPS & 69 & 11 \\
  & ResNet50 &  180 & 88 & 16 \\
  & CLIP-ViT-B-32  & 92 & 59 & 24 \\
  \midrule
  \multirow{3}{*}{\rotatebox{90}{BS 8}} & MobileViT  & 158 & 87 & 11 \\
  & ResNet50 &  223  & 117 & 19 \\
  & CLIP-ViT-B-32  & 101 & 68 & 27  \\
  \bottomrule
\end{tabular}
\end{table}
\section{Results}
\textbf{Low-shot and fine-tuned models approach expert-level gaze classification.} We assessed the CLIP-ViT-B-32 model's performance in a zero-shot setting, in which the model was not trained on the infant resuscitation dataset, across multiple input types. Using Top-1 and Top-3 accuracy metrics, we found that inputting only the raw frame for classification led to a dismal Top-1 accuracy of 8.96\% and a Top-3 accuracy of 38.39\%. However, performance increased when we used cropped images at a 128-pixel radius around the Tobii gaze estimate (Crop\textsubscript{128}), achieving a Top-1 accuracy of 36.93\% and a top-3 of 62.22\%. Likewise, Crop\textsubscript{256} exhibited a similar trend, with a slight uptick in Top-1 to 37.92\%, although with a decline in Top-3 accuracy to 49.39\%. Incorporating the object segmentation masks, either alone or in conjunction with cropping, dramatically increased accuracy; specifically, the “Crop\textsubscript{128} + Mask” configuration reached an impressive Top-3 accuracy of 76.10\%. Consequently, the joint input of cropping and segmentation masking yielded the most promising zero-shot gaze classification. During the few-shot learning phase, we assessed the performance of tip-adapted CLIP \citep{tip_adapter} on the test set using either the cropped image or the crop-mask pair as input. Remarkably, we found that with only 16 "featured" (no training involved) images, the feature adapter boosted CLIP's Top-1 accuracy to 71.17\% and Top-3 accuracy to 91.67\%.

We next fine-tuned a ResNet50, MobileViT, and CLIP-ViT on the EDIR image dataset under the “Crop\textsubscript{128} + Mask” resolution setting with limited training data. In the single-label scenario, where each image has only one ground truth label, MobileViT outperformed the other two models, achieving a Top-1 accuracy of 93.02\% and a Top-3 accuracy of 98.74\%. In comparison, CLIP-ViT-B-32 and ResNet50 yielded Top-1 accuracies of 87.44\% and 81.60\%, respectively. In the multi-label case, where multiple ground truth labels are possible, MobileViT again excelled, registering an mAP of 96.71\% and an F1-score of 91.60\%. This finding underscores the model's adeptness at learning from both cropped images and segmentation masks even for myriad ground truth labels. CLIP-ViT-B-32 and ResNet50 followed with mAPs of 92.39\% and 87.72\%, and F1-scores of 85.70\% and 77.68\%, respectively. Collectively, these findings endorse the utility of vision transformers for learning from a sparse dataset (e.g., 6 videos, each under 3 min.) to boost the accuracy and dependability of gaze analysis in neonatal care settings (see Fig. \ref{fig:fig2} for example predictions by MobileViT).

\begin{figure*}[tp]
    \centering
    \includegraphics[width=1\textwidth]{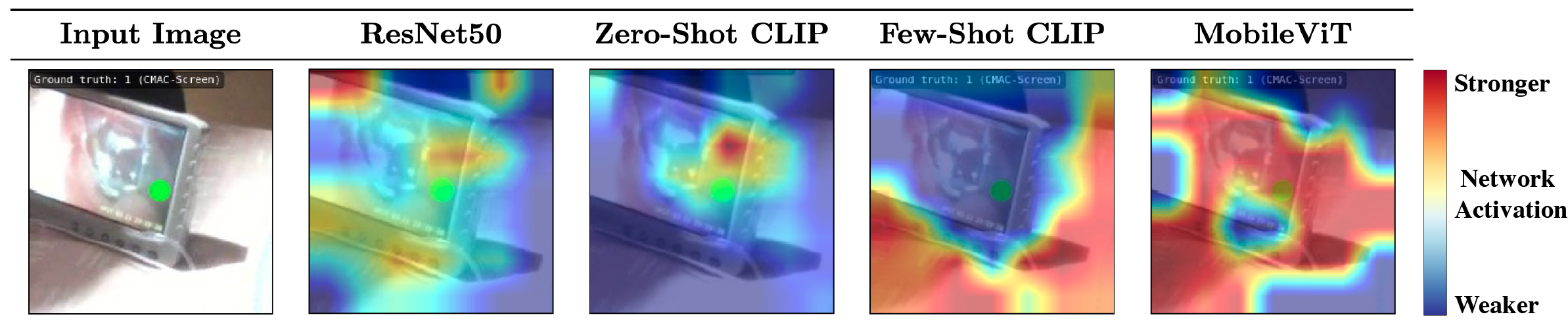}
    \caption{\textbf{Model class activation maps with GradCAM on the Laryngoscope Screen.} Each heat map conveys where the model is “looking” in this example image, where each model correctly predicted the class label. Less accurate models, like the ResNet-50, have more diffuse heat maps whereas the higher-performing fine-tuned CLIP (middle) and the MobileViT (far-right) models have heat maps concentrated on the object of attention, the Laryngoscope screen.}
    \label{fig:fig3}
\end{figure*}

\textbf{Class activation maps visualize the model's attention.} After model training, we employed grad-CAM, a technique for generating visual explanations in computer vision, to inspect the class activation maps (CAMs) for each model \citep{selvaraju_grad-cam_2020}, using testing images. In neural networks, "activation" reveals how specific input regions influence the model’s weights; CAMs pinpoint areas deemed crucial by the model for classification. Specifically, grad-CAM computes gradients of the target class score relative to feature map activations, resulting in a localization map that highlights vital regions for prediction. This map is superimposed on the input image, providing a visualization of the model's decision-making rationale (Fig. \ref{fig:fig3}). Notably, both MobileViT and the few-shot trained CLIP models showcased sharply focused heatmaps, signaling exceptional gaze classification precision. In contrast, the ResNet50 and zero-shot CLIP models produced more dispersed activation maps, reflecting diminished performance in this context.

\begin{figure*}[htbp]
    \centering
    \includegraphics[width=.9\textwidth]{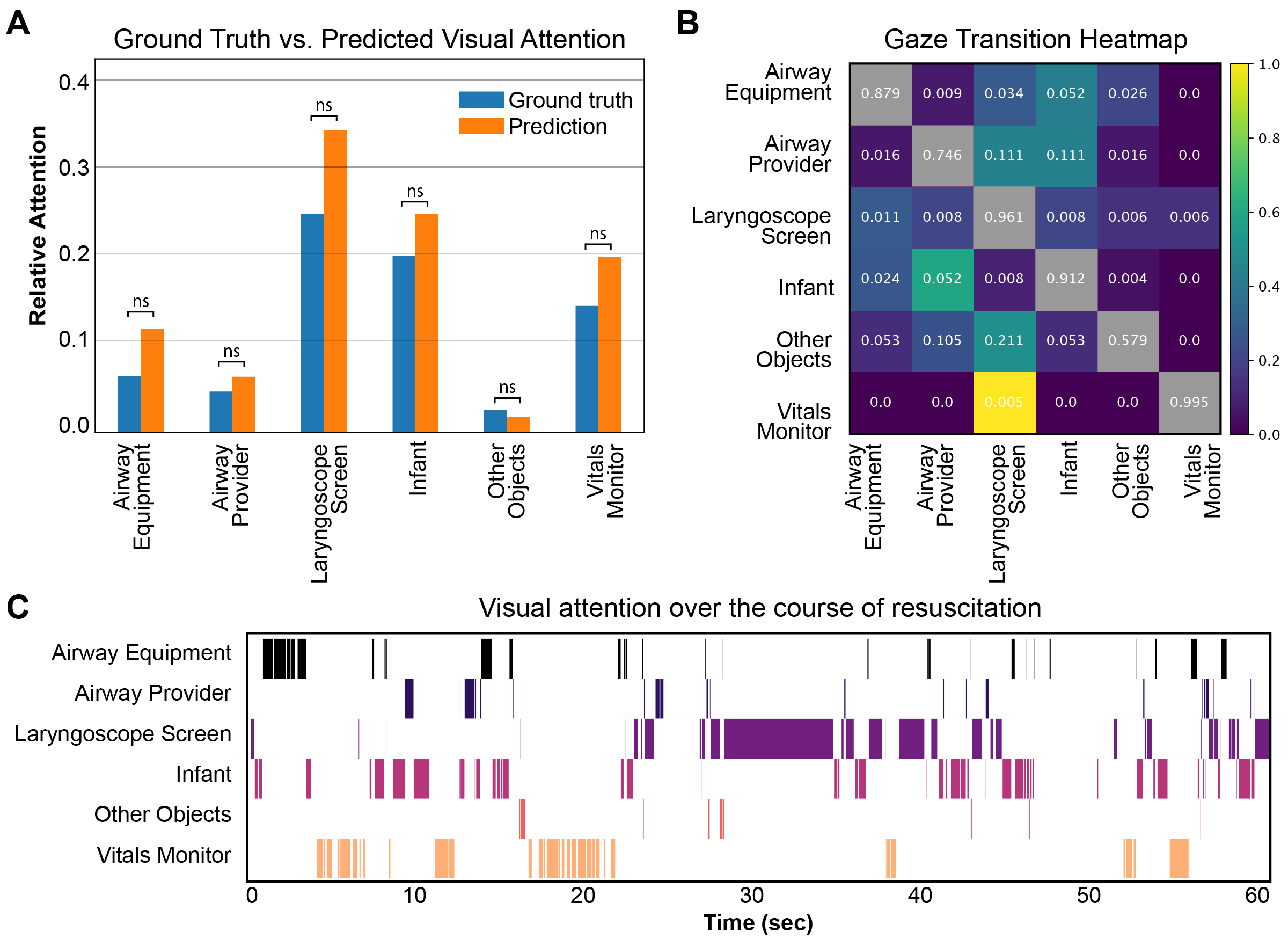}
    \caption{\textbf{Automated pipeline captures neonatologist gaze dynamics.} \textbf{(A)} Using the MobileViT model for semantic gaze classification, predicted visual attention approaches the expert-level annotations (ns: not significant --- i.e., our model makes predictions as accurately as our human annotators). \textbf{(B).} Visualizing gaze transitions between areas of interest. Transition matrix probabilities should be read as directed arrows from labels on the left to labels on the bottom, and are normalized per-row. Non-transitions (self-loops along the diagonal) are reported but omitted from the colormap. Notable findings include the strong \textit{Vitals $\rightarrow$ Laryngoscope Screen} transition, and the \textit{Infant $\leftrightarrow$ Airway-Provider} cycle, which may be due to their spatial proximity. \textbf{(C).} Visualizing gaze transitions over the course of a session. Notable findings include the blocks of \textit{Airway Equipment} early on (i.e., during placement), the \textit{Laryngoscope Screen} block (i.e., during intubation), and the \textit{Vitals Monitor} block (i.e., during patient stabilization).}
    \label{fig:fig4}
\end{figure*}

\textbf{Automated Pipeline Accurately Captures Neonatologist Gaze Dynamics.} We next evaluated the prediction capabilities of the best-performing model – the MobileViT – on single-label semantic gaze analysis. We first used the model to run inference on a held-out test video (Fellow\_56), whose ground-truth annotations we had. This model yielded a Top-1 accuracy of 95\% for this video. For each of the six predicted classes – six because there was no "Non-Team-Member" present in the video – we computed its relative frequency in the ground truth and predicted data. To assess whether the observed and expected frequencies were significantly different on a per-class basis, we performed a z-test for two proportions. Specifically, for each class, we evaluated whether the difference between the observed and expected proportions was statistically significant. We found that all but two of the classes were not significantly different from one another – that is, the predictions of our semantic gaze classification model were statistically equivalent to those in the ground truth dataset (p $>$ 0.05; z-statistic: ~0.299). The remaining two classes – "Airway Equipment" and "Other Physical Objects" were found to be significantly different. This could be a true result, given the overlap in frequency between airway equipment, infant, and airway provider and the arbitrary classification of "other physical objects". However, when controlling for multiple statistical tests with the Bonferroni correction, in which we divided the threshold for significance by the number of classes tested, we found that the predicted relative frequency of all classes was not statistically different from the ground truth (Figure \ref{fig:fig4}). This suggests that our pipeline can precisely and automatically classify semantic gaze from eye-tracking video alone. To visualize neonatologist visual attention, we computed the transition matrix between classes and plotted the gaze transitions, with scaled nodes and edges for those classes that had greater transitions. Finally, we plotted visual attention throughout resuscitation (Fig \ref{fig:fig4}c).

\textbf{Real-Time Semantic Gaze Classification Enables Clinical Integration.} To assess the computational efficiency of the machine learning models, we conducted inference speed experiments of our three models – ResNet-50, MobileViT, and CLIP-ViT-B-32 – across multiple hardware platforms. The tested hardware configurations included: NVIDIA RTX A6000 GPU; NVIDIA RTX 3080 GPU; AMD Ryzen Threadripper PRO 5975WX CPU; and 11th Gen Intel Core i7-11700K CPU. To capture the model’s adaptability to real-time and batch processing scenarios, each model was evaluated under two different batch sizes: a single-instance batch (BS=1) and a batch of eight instances (BS=8). For each batch size and hardware combination, inference speed was measured in frames per second (FPS). To mitigate the effects of background processes, no other significant tasks were executed concurrently. Additionally, the models were allowed a "warm-up" period to ensure that all components were operating at their peak capabilities. Following the completion of 10 repetitions per configuration, we computed the average FPS of each model, which is reported in Table \ref{tab:table1}. The ResNet50 model led in speed across most configurations, reaching 180.29 FPS and 88.35 FPS on the NVIDIA RTX A6000 and RTX 3080 GPUs, respectively, with single-instance batches (BS = 1). MobileViT also demonstrated efficiency, exceeding the 25 FPS frame rate of the Tobii eye-tracking glasses on three of the four hardware platforms tested. For instance, it clocked 138.39 FPS and 69.39 FPS on the NVIDIA RTX A6000 and RTX 3080 GPUs, respectively. Even with a batch size of eight (BS = 8), MobileViT maintained an average speed of 41.16 FPS on the AMD Ryzen Threadripper PRO 5975WX CPU. Expert human annotators in our study labeled frames at rates ranging from 0.5 to 5 FPS, an order of magnitude slower than even our least efficient models. This highlights the potential of our approach for initial classification tasks. While ResNet50 may boast the highest raw speed, MobileViT offers a balanced profile of speed and adaptability across diverse hardware, making it ideally suited for real-time decision support in clinical settings. Our system is thus well-equipped for real-time semantic gaze classification and visual attention quantification.

\section{Discussion}
\textbf{Conclusions.} Here, we report significant progress in semantic gaze classification in the context of eye-tracking during neonatal resuscitation. Unlike prior work which primarily focused on simulation-based studies or the educational applications of eye-tracking \citep{katz_visual_2019, leone_using_2019}, our approach achieves highly accurate ($>$93\% Top-1 accuracy) automated analysis of provider gaze patterns during live neonatal resuscitation scenarios, offering an improvement over the traditional manual post-hoc area of interest analysis \citep{weinberg_visual_2020}. Further, our approach enables realistic real-time applications. Utilizing lightweight, cutting-edge deep learning models, our pipeline addresses the limitations associated with both simulated environments and data-intensive approaches, opening new avenues for immediate and context-relevant analysis in critical care scenarios and beyond.

\textbf{Limitations.} Despite promising findings, translating our results into real-time clinical applications faces challenges. Managing eye-tracking device calibration and video recording without disrupting medical tasks is crucial. Privacy concerns demand strict data security for patient video data. Our reliance on eye-tracking calibration \citep{browning_use_2016} may suffer from issues like suboptimal glasses placement and environmental factors. Moreover, deep learning models lack transparency, underscoring the need for model explainability in real-world environments \citep{ma_eye-gaze-guided_2022}. Lastly, validating our approach across various settings, scenarios, and tasks at multiple sites is essential for its broader applicability.

\textbf{Impact.} Our system has important implications within neonatology as well as the medical field at large, as it can inform specialized training programs for providers, thereby improving attention management and decision-making during high acuity settings. These insights could also aid in optimizing delivery room and NICU layouts for better workflow and focus distribution among patients, monitors, and medical teams. Additionally, our research lays the groundwork for further exploration of attention dynamics in critical care to better understand and potentially reduce cognitive load.

\textbf{Future Directions.} We will expand our dataset to include a wider range of hospitals and clinical settings within neonatology to improve the generalizability of our models. We plan to validate our system's effectiveness in real-world clinical settings, augmenting the dataset with simulated data for greater robustness. We aim to apply our approach in various medical domains, such as radiology and surgery, and to additional eye-tracking systems, such as Pupil Labs. Our long-term vision is to create a comprehensive monitoring system that combines gaze data with other sensor-based metrics to assess provider workload, fatigue, situational awareness, and decision-making holistically. We are also exploring additional metrics such as whole-body pose estimation of providers during intubation \citep{pose_health}, comparative studies in high-risk delivery rooms \citep{batey_newborn_2022}, and correlating eye-tracking metrics with biometric data to enhance our understanding of attention management in critical care scenarios \citep{herrick_impact_2020}, striving for a more precise and data-driven approach.
\section{Acknowledgments}
We thank the healthcare providers of the Children's Hospital of Philadelphia (CHOP) for their support and collaboration. This project was supported by the American Academy of Pediatrics Neonatal Resuscitation Program Human Factors Grant (awarded to HMH). HMH is supported by the Agency for Healthcare Research and Quality career development grant (K08HS029029). MLP is supported by R37-MH109728, R01-MH108627, R01-MH-118203, KA2019-105548, U01MH121260, UM1MH130981, R56MH122819, R56AG071023.
{
    \small
    \bibliographystyle{ieeenat_fullname}
    \bibliography{main}
}


\end{document}